\documentclass[]{IEEEtran}
\usepackage{cite}
\usepackage{amsmath,amssymb,amsfonts}
\usepackage{algorithmic}
\usepackage{graphicx}
\usepackage{textcomp}
\usepackage{xcolor}
\usepackage{multirow}
\usepackage{caption} 
\usepackage{float}
\usepackage{soul}
\usepackage{xcolor}
\usepackage{amsmath}
\usepackage{hyperref}
\newcommand{\etal}{\textit{et al.}}

\usepackage{array}
\newcommand{\PreserveBackslash}[1]{\let\temp=\\#1\let\\=\temp}
\newcolumntype{C}[1]{>{\PreserveBackslash\centering}p{#1}}
\newcolumntype{R}[1]{>{\PreserveBackslash\raggedleft}p{#1}}
\newcolumntype{L}[1]{>{\PreserveBackslash\raggedright}p{#1}}
\usepackage{booktabs}

\usepackage{subcaption}
\hypersetup{
    colorlinks=true,
    allcolors=blue
}

\def\BibTeX{{\rm B\kern-.05em{\sc i\kern-.025em b}\kern-.08em
    T\kern-.1667em\lower.7ex\hbox{E}\kern-.125emX}}
\graphicspath{ {images/} }

\begin{document}

\title{Rethinking Cooking State Recognition with Vision Transformers} 

\author{ 
    \IEEEauthorblockN{Akib Mohammed Khan\textsuperscript{*}, Alif Ashrafee\textsuperscript{*}, Reeshoon Sayera\textsuperscript{*}, Shahriar Ivan, and Sabbir Ahmed\\}

    \IEEEauthorblockA{Department of Computer Science and Engineering,\\Islamic University of Technology, Gazipur, Bangladesh\\}
    
    \IEEEauthorblockA{Email: \{akibmohammed, alifashrafee, reeshoonsayera, shahriarivan, sabbirahmed\}@iut-dhaka.edu}
}


\IEEEpubid{
\begin{minipage}[t]{\textwidth}\ \\[10pt]
      \small{* Contributed Equally\\
      (Copyright $\copyright$ 2022 IEEE. This work has been submitted to the IEEE for possible publication. Copyright may be transferred without notice, after which this version may no longer be accessible.)}
\end{minipage}
}

\maketitle

\begin{abstract}

To ensure proper knowledge representation of the kitchen environment, it is vital for kitchen robots to recognize the states of the food items that are being cooked. Although the domain of object detection and recognition has been extensively studied, the task of object state classification has remained relatively unexplored. The high intra-class similarity of ingredients during different states of cooking makes the task even more challenging. Researchers have proposed adopting Deep Learning based strategies in recent times, however, they are yet to achieve high performance. In this study, we utilized the self-attention mechanism of the Vision Transformer (ViT) architecture for the Cooking State Recognition task. The proposed approach encapsulates the globally salient features from images, while also exploiting the weights learned from a larger dataset. This global attention allows the model to withstand the similarities between samples of different cooking objects, while the employment of transfer learning helps to overcome the lack of inductive bias by utilizing pretrained weights. To improve recognition accuracy, several augmentation techniques have been employed as well. Evaluation of our proposed framework on the `Cooking State Recognition Challenge Dataset' has achieved an accuracy of 94.3\%, which significantly outperforms the state-of-the-art.


\end{abstract}

\begin{IEEEkeywords}
Kitchen robots, Object state classification, Vision Transformers, Cooking ingredients recognition, Food classification
\end{IEEEkeywords}

\section{Introduction}\label{sec:introduction}
Recent revolutions in the field of robotics have led to the development of various types of service robots, and more specifically, kitchen robots. Such autonomous systems have to deal with several phases of cooking a meal where the shape and state of ingredients vary depending on the food that is prepared. Recognizing these states is rather straightforward for a professional chef but challenging for a robot. A robust framework with the ability to accurately classify the states of different ingredients can be extremely useful for such robots to ensure the proper handling of the items \cite{Fonseca2019kitchenRobots}. 

Although there has been a substantial amount of study in the field of object detection and recognition, the domain of object state recognition has received comparatively little attention. More specifically, food state recognition is an even more obscure field of study. However, this is a rather interesting task, which finds application in various research domains such as automatic video recipe transcript, service automation, fine-grained cooking activity understanding, robot task planning and manipulation control, and so on~\cite{ciocca2020state,berezina2019robots}. Moreover, kitchen robots can provide a great improvement in the quality of life, especially for the elderly and disabled people \cite{ma_yan_fu_zhao_2011}.

Cooking state recognition is an objectively difficult task for an autonomous system as the same food can have different appearances due to the preparations, lighting conditions, placements, and angle of image acquisition \cite{ciocca2020state}. Earlier works exploited hand-crafted visual features that are needed to be carefully selected based on the task, and do not generalize well on samples with high diversity \cite{bosch2011combining, hoashi2010image}. However, the advent of Convolution Neural Network (CNN)-based architectures has produced a remarkable performance with their automatic feature learning capability and have been widely used in various recognition and classification tasks \cite{li2021aSurvey, ahmed2022lessIsMore, yasmeen2021csvcNet, Ashrafee_2022_WACV}. Moreover, a more recent improvement of these architectures shows that the self-attention mechanism can achieve even higher performance in image classification tasks with their ability to learn globally salient features \cite{khan2022transformers}.  

In this paper, we proposed a method for the Cooking State Recognition task using Vision Transformers (ViT) utilizing global attention and adding positional embeddings to each patch of the image so that it can retain its spatial information which is lost when using traditional neural networks. Being pretrained and fine-tuned respectively on the ImageNet and Cooking State Recognition Challenge datasets, the model was able to distinguish among different objects having similar states. The performance of the proposed approach was compared with the several state-of-the-art Deep CNN architectures and the existing works in this domain, where it significantly outperformed all of them.

The rest of the paper is laid out as follows. We discussed the related works in Section \ref{sec:lit_review}. Section \ref{sec:methodology} discussed the different components of the proposed pipeline. The experimental findings are described in Section \ref{sec:results}. Finally, Section \ref{sec:conclusion} contains the concluding remarks along with some future directions.

\begin{figure*}[t]
    \centering
    \includegraphics[width=0.95\textwidth]{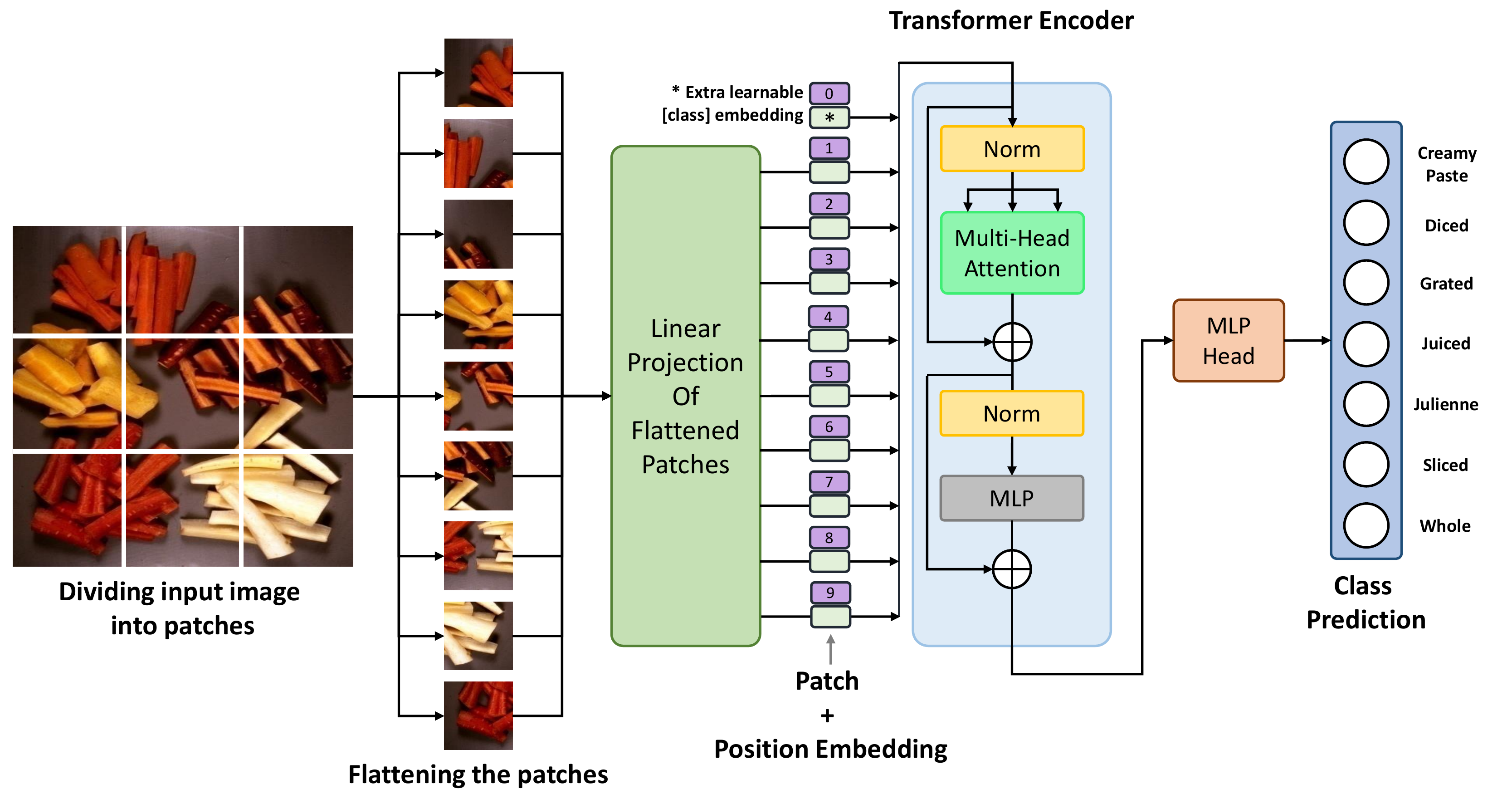}
    \caption{Overview of the Vision Transformer architecture (Adapted from \cite{vit})}
    \label{archi}
\end{figure*}

\section{Literature Review}\label{sec:lit_review}

There have been recent works on interpreting cooking recipes \cite{malmaud2015s}, food image recognition \cite{hassannejad2016foodImage}, prediction of cooking tasks \cite{chen2018deepUnderstanding}, etc. However, the classification of the states of cooking ingredients for a kitchen robot is a relatively newer domain to investigate. Ciocca \textit{et al.} showed that the problem of state recognition is more approachable through the use of automatically learned features as they generally outperform hand-crafted features by a large margin \cite{ciocca2020state}. Specific to the problem at hand,  researchers used modified and/or ensembled versions of CNN-based architectures for the classification state of cooking objects \cite{salekin2019cooking, paul2018classifyingCooking, chen2018identifying}.

In \cite{salekin2019cooking}, the authors proposed a modified version of the InceptionV3 model \cite{inception} where they added two more convolutional layers and a Global Average Pooling layer at the end and achieved an accuracy of 73.3\%. The authors in \cite{paul2018classifyingCooking} utilized a pretrained VGG16 model \cite{vgg} while \cite{chen2018identifying} proposed a few modifications to the VGG19 architecture by tuning the filter sizes of different layers. A similar approach was followed in \cite{wilches2019VGG}, where a VGG network with 19 layers was pretrained on the ImageNet dataset and then fine-tuned on a dataset of cooking state recognition containing 10 classes. However, most of these works have failed to distinguish samples of the states having high inter-class similarity, which resulted in poor recognition accuracy. In this regard, the recently emerged models such as Vision Transformers (ViT) \cite{vit} can be utilized which exploits a transformer architecture by adding positional embeddings to patches of an image and applying self-attention so that the images can retain their positional information which gets lost when using traditional neural networks.

\section{Methodology}\label{sec:methodology}

\subsection{Model Description}

Vision transformer (ViT) is a model based on the architecture of natural language transformers that uses a multi-headed self-attention mechanism on images for classification purposes. While the standard transformation network used in Natural Language Processing (NLP) takes a 1D sequence of embedding vectors as input \cite{attention}, ViTs deal with images, hence the input needs to be converted into a 1D sequence. This is done by splitting the images into patches, flattening the patches, and then producing lower-dimensional linear embeddings from those patches. 1D position embeddings are appended to the patch sequence to provide positional information. The sequence of patches and the position embeddings from the input is then fed to the encoder \cite{vit}.

The architecture of the transformer encoder, as shown in the \figureautorefname~\ref{archi}, is made of alternating layers of Multi-Headed Self-Attention (MSA) and Multi-Layer Perceptron (MLP). Each MSA and MLP block has a normalization layer applied before it, and residual connections applied after it. Each MLP has two layers with a GELU activation function. The core part of ViTs is the MSA blocks, which dynamically compute weights for each image patch embedding. The embedding vectors of these image patches are aggregated by a weighted sum in which the weight of each embedding is given by the attention scores. As a result, we get a weighted average of all the image vectors representing the patches, enabling the model to integrate information across the entire image, as compared to CNNs, where the kernels used are small windows that encompass a small spatial range in the image.

\begin{figure*}[ht]
    \centering
        \begin{subfigure}[c]{0.13\textwidth}
            \centering
            \includegraphics[width=1\textwidth]{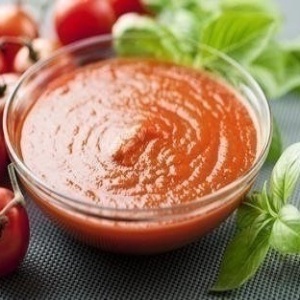}
            \label{im1}
            \caption{Creamy Paste}
        \end{subfigure}
        \begin{subfigure}[c]{0.13\textwidth}
            \centering
            \includegraphics[width=1\textwidth]{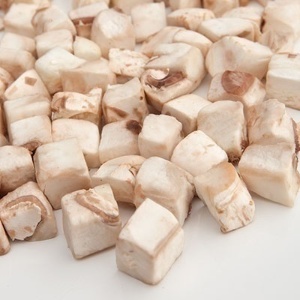}
            \label{im2}
            \caption{Diced}
        \end{subfigure}
        \begin{subfigure}[c]{0.13\textwidth}
            \centering
            \includegraphics[width=1\textwidth]{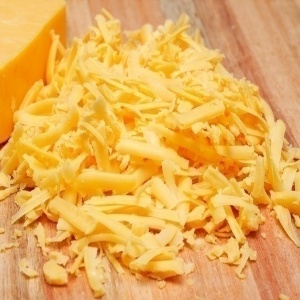}
            \label{im3}
            \caption{Grated}
        \end{subfigure}
        \begin{subfigure}[c]{0.13\textwidth}
            \centering
            \includegraphics[width=1\textwidth]{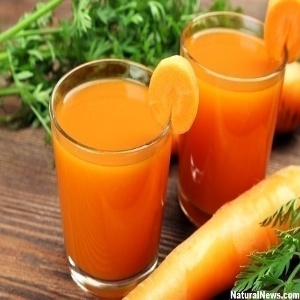}
            \label{im4}
            \caption{Juiced}
        \end{subfigure}
        \begin{subfigure}[c]{0.13\textwidth}
            \centering
            \includegraphics[width=1\textwidth]{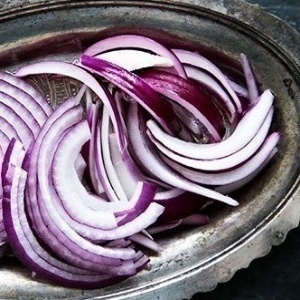}
            \label{im5}
            \caption{Jullienne}
        \end{subfigure}
        \begin{subfigure}[c]{0.13\textwidth}
            \centering
            \includegraphics[width=1\textwidth]{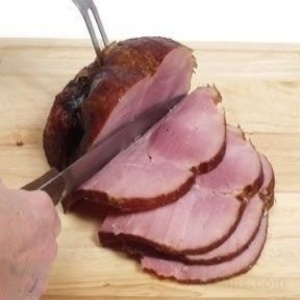}
            \label{im6}
            \caption{Sliced}
        \end{subfigure}
        \begin{subfigure}[c]{0.13\textwidth}
            \centering
            \includegraphics[width=1\textwidth]{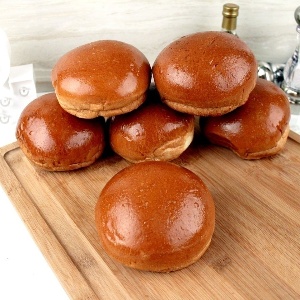}
            \label{im7}
            \caption{Whole}
        \end{subfigure}
        \caption{Samples from the Cooking State Recognition Challenge dataset}
        \label{dataset-images}
\end{figure*}

\subsection{Inductive Bias}
Inductive bias refers to the set of assumptions a model uses to perform predictions on unseen data. Vision transformers have much less image-related inductive bias compared to CNNs. A convolution operation takes into account locality, 2D neighborhood structure, and translation equivariance. ViTs use the two-dimensional local neighborhood information very sparingly as the self-attention layers are global, while only the MLP layers have locality and translation equivariance. The position embeddings carry no 2D information, so the ViT has to learn all spatial relations between all the image patches from scratch, enabling it to learn better from data than CNNs.

\subsection{Transfer Learning}

Transfer learning is a technique where the knowledge a model has learned from one task can be reused in another task \cite{ashikur2022twoDecades}. When a pretrained model is used as the starting point in a classification task, training time speeds up, and the performance increases \cite{morshed2022fruit}. The lack of inductive biases in ViT does not allow it to generalize well when trained on smaller datasets. This is where transfer learning becomes a better-suited approach while trying to apply ViT on insufficient data. Researchers have found that when ViT is pretrained on a large-scale dataset like ImageNet-21k \cite{imagenet}, and the weights are transferred during fine-tuning with smaller datasets, it has produced outstanding results outperforming multiple state-of-the-art image recognition benchmarks \cite{raghu2021neurips}. We have used the ViT model pretrained on the ImageNet-21k dataset.

\section{Results and Discussion}\label{sec:results}

\subsection{Dataset}
We used the `Cooking State Recognition Challenge dataset' proposed by Jelodar \etal~\cite{jelodar2018identifyingObject}. The dataset consists of 5902 images of 7 cooking states from 18 types of objects. The cooking states along with their count are diced (700), julienne (672), whole  (1304), juiced (638), creamy paste (730), grated (819), and sliced (1315). \figureautorefname~\ref{dataset-images} shows samples of images from different states from the dataset. The `Whole' state contains objects in their original form such as a whole tomato or a whole onion. Objects in chopped-up or cut-up forms, such as diced onions or cubed chicken are classified as the `Diced' state. The `Julienne' state contains ingredients cut in thin strips, like that of carrots. The `Juiced' state represents objects in liquid form such as tomato juice or milk.  The `Creamy paste' contains objects that have been mashed to form paste having a creamy like texture, such as mashed potatoes or cheese. The `Grated' state represents the ingredients that have been reduced to shreds such as grated carrot or grated cheese. Objects that have been thinly cut into smaller portions, like sliced bread, etc, belong to the `Sliced' state. 

All images were resized to $224\times224$ dimension. Normalization and sample-wise centering were performed to ensure zero mean and unit standard deviation. At first, 85\% of the samples were used in the training set and the rest 15\% in the test set. Furthermore, 15\% of the training set data was used as the validation set. In summary, 4106 images were used in training, 728 images in validation, and 1068 images in testing.

\subsection{Experimental Setup}
The experiments were conducted on the Kaggle platform using Tesla V100 GPU. Two variants of the ViT models were used, namely the Base-16 (ViT B-16) and Large-16 (ViT L-16), where the input was divided in $16\times16$ patches. The configuration of each variant is shown in \tableautorefname~\ref{vitConfig} and has been followed as per the recommendation of \cite{vit}.

\begin{table}[t]
\centering
\caption{Configuration of ViT model variants}

    \begin{tabular}{C{1.2cm} C{1cm} C{1cm} C{0.6cm} C{0.6cm} C{1.5cm}}
    \toprule
    \textbf{Model} & \textbf{Layers} & \textbf{Hidden size D} 
    & \textbf{MLP size} & \textbf{Heads} & \textbf{Parameters Count (Millions)} \\ \midrule 
    ViT B-16 & 12 & 768 & 3072 & 12 & 86 \\ 
    ViT L-16 & 24 & 1024 & 4096 & 16 & 307 \\ 
    \midrule   
\end{tabular}
\label{vitConfig}
\end{table}

We have utilized the Cross-Entropy loss function which is suitable for a multi-class classification task. The model was trained for 10,000 steps with Early Stopping callback and a batch size of 32. A Learning Rate (LR) of 0.03  was used as a higher LR is generally suitable for small datasets and a Cosine decay was applied for learning rate scheduling. We used the stochastic gradient descent (SGD) optimizer as it provided the best results on the Cooking State Recognition Challenge \cite{salekin2019cooking}. Random rotation, horizontal flip, conversion to Hue Saturation Value (HSV) form, random brightness-contrast, and random shift-scale were among the data augmentation techniques we applied to the training set to increase the size by five-fold using the augmentation tool `Albumentations' \cite{album}.






\subsection{Performance of different baseline architectures}

For years, CNN-based architectures have been dominant in image classification tasks and only recently transformer-based architectures have emerged and taken over these networks \cite{khan2022transformers}. We evaluated several such state-of-the-art deep CNN architectures on the Cooking State Recognition Challenge dataset using weights\footnote{\url{https://keras.io/api/applications/}} pretrained on the ImageNet dataset  \cite{imagenet} 
and compared the results with the transformer-based architecture of the ViT B-16 model. For this experiment, we chose these models due to their superior performance in image classification tasks in recent times. From the results in \tableautorefname~\ref{cnnvsvit}, we can observe that among the CNN-based models, MobileNet \cite{mobilenet} performs the best while having the least amount of parameters. However, in general, the learning of these models have a tendency to plateau after reaching a certain accuracy which is unsatisfactory. This is apparent when the ViT model achieves an accuracy of 93\% which is a huge increase compared to that of MobileNet which has an accuracy of 79\%. It is to be noted that ViT has a significantly larger parameter count, however, they are relatively cheaper to train on smaller datasets with the help of transfer learning. Moreover, the sheer improvement in accuracy is of paramount importance which makes the ViT a marked upgrade over the traditional deep CNN models.

\begin{table}[t]
\centering
\caption{Performance comparison of the baseline pretrained architectures}
\label{cnnvsvit}
\begin{tabular}{L{2.5cm} C{1.3cm} C{2.5cm}}
\toprule
\textbf{Architecture} & \textbf{Accuracy (\%)} & \textbf{Parameter count (millions)} \\  
\midrule
ResNet50 \cite{resnet} & 63.0 & 25.6   \\ 
ResNet152 \cite{resnet} & 66.0 & 60.4 \\ 
DenseNet169 \cite{densenet} & 71.0 & 14.3 \\ 
DenseNet201 \cite{densenet} & 75.0 & 20.2  \\ 
InceptionV3 \cite{inception} & 75.0 & 23.9  \\
MobileNet \cite{mobilenet} & 79.0 & 4.30  \\ 
\midrule
Ours & \textbf{93.0} & 86.0 \\  \bottomrule
\end{tabular}
\end{table}

\subsection{Ablation Study}

We have conducted experiments incorporating different modules of the proposed pipeline to determine which combination works best and the results of each experiment are outlined in \tableautorefname~\ref{ablation}. The three different variables of our ablation study include the choice of the model (ViT B-16 vs ViT L-16), state of weights (scratch vs pretrained), and augmentation (original vs augmented dataset). It is evident that using a pretrained model on the ImageNet dataset drastically outperforms the ViT model trained from scratch. As discussed earlier, this aligns with the fact that applying transfer learning for ViT models on small datasets produces much better results compared to training from scratch.

\begin{table}[b]
\centering
\caption{Ablation study of different components of the proposed pipeline}
\label{ablation}
\begin{tabular}{c c c c}
\toprule
  \textbf{Model} &
  \textbf{Pretrained} &
  \textbf{Augmented} &
  \textbf{Accuracy (\%)} \\ 
  
\midrule
ViT B-16 & $\times$ & $\times$  & 55.4 \\ 
ViT B-16 & $\times$ & \checkmark  & 55.1 \\ 
ViT B-16 & \checkmark & $\times$  & 93.0 \\ 
ViT B-16 & \checkmark & \checkmark  & 94.3 \\ 
ViT L-16 & \checkmark & $\times$  & 94.0 \\ 
ViT L-16 & \checkmark & \checkmark  & 95.0 \\ 
\midrule
\end{tabular}
\end{table}


From the table, we can see that using augmentations alone does not affect the results when ViT is trained from scratch as the data is still insufficient for the model to generalize due to the lack of inductive biases. However, augmenting the dataset improves the results for both the pretrained variations of ViT. The increase in training set size along with leveraging weights of a larger dataset allows the model to generalize well on unseen data. The final ablation was to evaluate the different variants of ViTs available, namely the Base-16 model with fewer parameters and its larger counterpart Large-16 model with significantly higher parameters. \tableautorefname~\ref{ablation} shows that even though the larger model demonstrates a slight improvement in accuracy (0.7\%), it is at the expense of 221 million additional parameters which is a diminishing trade-off in terms of computational complexity. Moreover, the large model has a higher requirement in terms of training and inference time. The similarity in results, the time and computational overhead of the large model lead us to the conclusion that the base model is a better choice when using pretrained weights on the augmented dataset.

\subsection{Class-wise Analysis}
\tableautorefname~\ref{classificationReport} summarizes the class-wise performance of the proposed pipeline. It can be observed that our model achieves the highest recall value of 97\% for the state diced as this class was most distinct in structure compared to others. The lowest recall value 92\% accounted for the states grated, jullienne and whole as they had the most similarity between samples of different cooking objects. The recall values for all other states are equal to or above 95\%. The average precision, recall, and F1 score are 0.94, 0.94, and 0.94 respectively. This signifies that the model learned better image representation by contextualizing the information globally resulting in outstanding performance.

\begin{table}[t]
\center
\caption{Per class precision, recall, and F1-score on the test split}
\begin{tabular}{l C{1cm} c c c}
\toprule
\textbf{Class} & \textbf{Sample Count} & \textbf{Precision} & \textbf{Recall} & \textbf{F-1 Score}\\ \midrule
Creamy    & 124 & 0.91 & 0.96 & 0.93 \\ 
Diced     & 144 & 0.93 & 0.97 & 0.95\\ 
Grated    & 131 & 0.93 & 0.92 & 0.93 \\ 
Juiced    & 147 & 0.95 & 0.96 & 0.96 \\ 
Jullienne & 110 & 0.94 & 0.92 & 0.93 \\ 
Sliced    & 237 & 0.95 & 0.95 & 0.95 \\ 
Whole     & 175 & 0.97 & 0.92 & 0.94 \\ \midrule
Average  &  \textbf{1068} & \textbf{0.94}      & \textbf{0.94}   & \textbf{0.94}      \\ \bottomrule
\end{tabular}
\label{classificationReport}
\end{table}


\begin{figure}[b]
    \centering
    \includegraphics[width=0.43\textwidth]{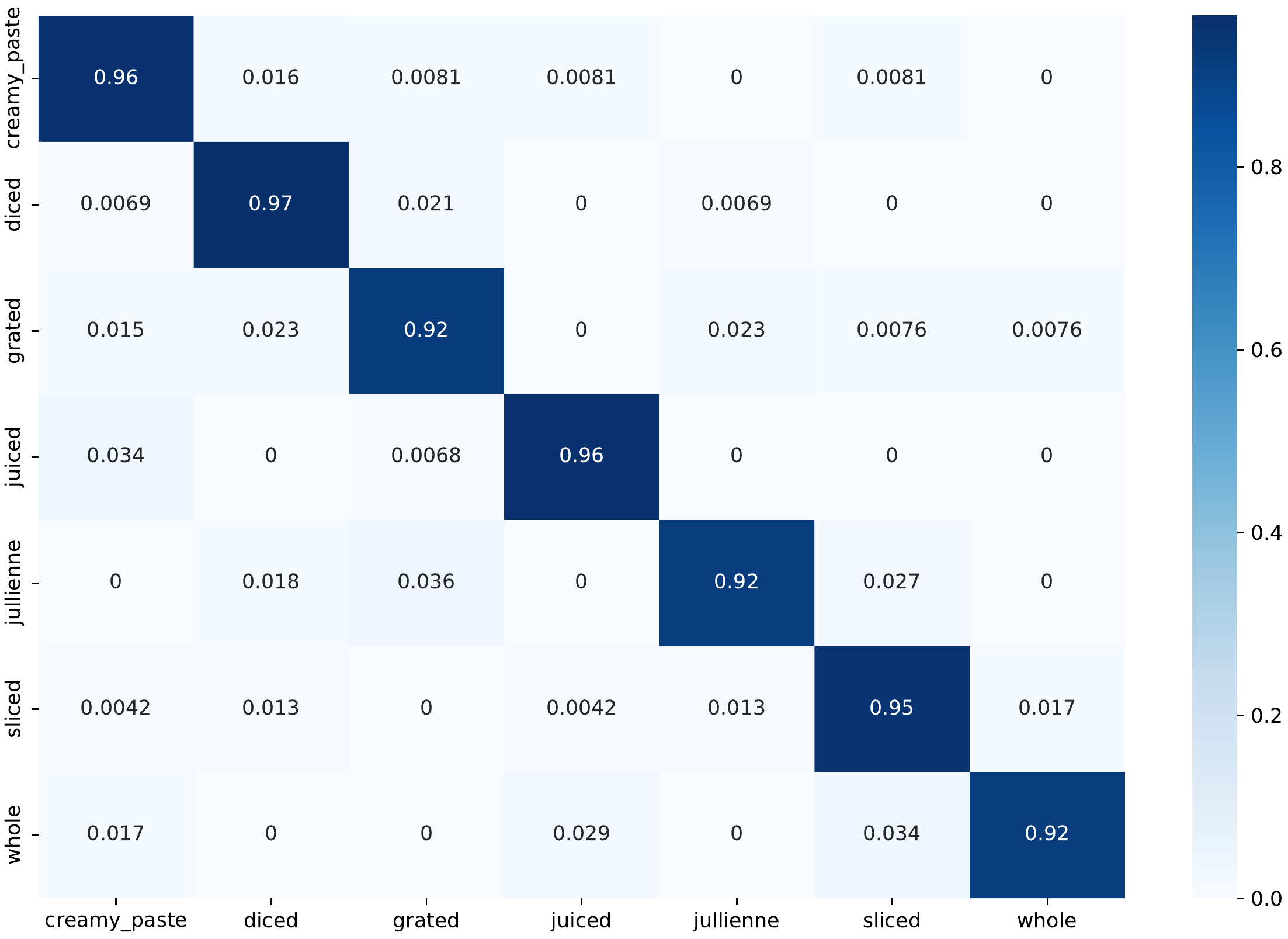}
    \caption{Normalized Confusion Matrix on the test dataset}
    \label{confusion}
\end{figure}

\begin{figure}[t]
    \centering
    \captionsetup[subfigure]{justification=centering}
        \begin{subfigure}[c]{0.24\textwidth}
            \centering
            \includegraphics[width=0.9\textwidth]{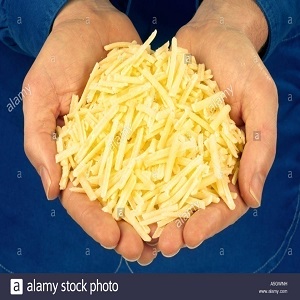}
            \label{qual1}
             \caption{Label: Grated}
        \end{subfigure}
        \begin{subfigure}[c]{0.24\textwidth}
            \centering
            \includegraphics[width=0.9\textwidth]{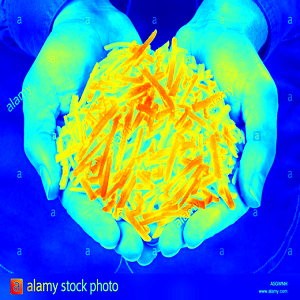}
            \label{qual2}
            \caption{Predicted: Grated}
        \end{subfigure}
        \begin{subfigure}[t]{0.24\textwidth}
            \centering
            \includegraphics[width=0.9\textwidth]{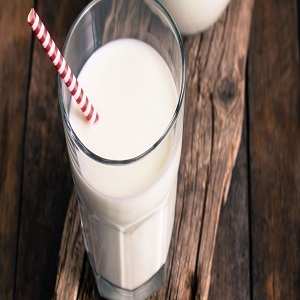}
            \label{qual3}
            \caption{Label: Juiced}
        \end{subfigure}
        \begin{subfigure}[t]{0.24\textwidth}
            \centering
            \includegraphics[width=0.9\textwidth]{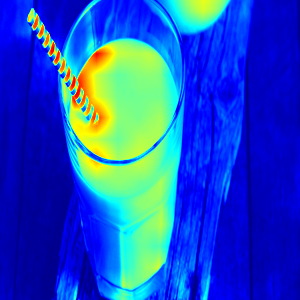}
            \label{qual4}
            \caption{Predicted: Juiced}
        \end{subfigure}
        \begin{subfigure}[t]{0.24\textwidth}
            \centering
            \includegraphics[width=0.9\textwidth]{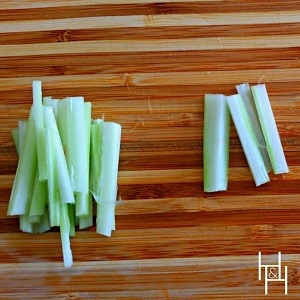}
            \label{qual5}
            \caption{Label: Julienne}
        \end{subfigure}
        \begin{subfigure}[t]{0.24\textwidth}
            \centering
            \includegraphics[width=0.9\textwidth]{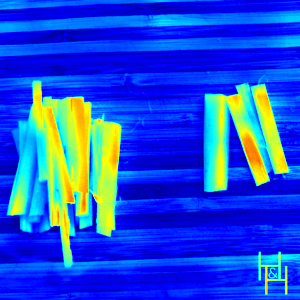}
            \label{qual6}
            \caption{Predicted: Julienne}
        \end{subfigure}
        \caption{Attention map output for correctly classified samples}
        \label{qualitative}
\end{figure}


\begin{table}[b]
\center
\caption{Performance comparison with the state-of-the-art works on the Cooking State Challenge dataset}
\begin{tabular}{l c c c}

\toprule
\textbf{Model}  & \textbf{Precision (\%)} & \textbf{Recall (\%)} & \textbf{Accuracy (\%)} \\ 
\midrule

Salekin \etal \cite{salekin2019cooking} & 71.0 & 70.0 & 73.0 \\
Paul \etal \cite{paul2018classifyingCooking}     & -  & - & 77.0 \\
Chen \etal \cite{chen2018identifying}    & - & - & 80.0 \\ 
\midrule
Ours & \textbf{94.0}      & \textbf{94.0}   & \textbf{94.3}     \\ \bottomrule

\end{tabular}
\label{comparesota}
\end{table}

\subsection{Comparison with state-of-the-art methods}
Recent works have proposed a few modified versions of Deep CNNs along with fine-tuning weights and hyperparameters for the cooking state recognition task \cite{salekin2019cooking, paul2018classifyingCooking, chen2018identifying}. However, due to the intra-class variability in the dataset of different object types having the same state, and the high inductive bias of CNNs in general, the results of these models tend to plateau in performance. As shown in Table \ref{comparesota}, Chen \etal \cite{chen2018identifying} achieves the highest accuracy (80\%) among the CNN-based methods. In contrast, the self-attention mechanism of ViTs helps withstand the intra-class variation, and leveraging the weights pretrained on the ImageNet trumps the lack of inductive bias, making it a formidable method for cooking state recognition as it exhibits a drastic improvement in results across all metrics compared to the modified CNN based methods.

\subsection{Qualitative Analysis}
In order to determine whether the proposed pipeline is learning to predict accurately while also considering the relevant features, we generated the attention map output for different samples. This visualization exhibits how the model highlights the semantically relevant regions for classification. These attention maps were produced by averaging the attention weights across all the heads, adding an identity matrix, and finally recursively multiplying the weight matrices across every layer  \cite{abnar-zuidema-2020-quantifying}. \figureautorefname~\ref{qualitative} shows some of the correctly classified examples of the attention map from the output token superimposed over the input images in the form of a heat map. The visualizations show that using the global self-attention mechanism allows the model to focus on relevant regions that conduce to correct predictions while suppressing irrelevant information that might cause the model to misclassify. Hence, the model understands where and what to look for in an image to make the correct class predictions, irrespective of the object type.

\begin{figure}[t]
    \centering

    \captionsetup[subfigure]{justification=centering}
        \begin{subfigure}[c]{0.24\textwidth}
            \centering
            \includegraphics[width=0.9\textwidth]{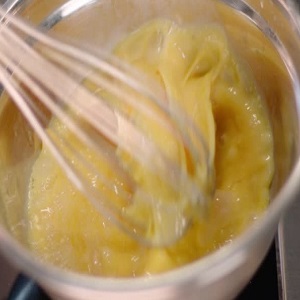}
            \caption{Label: Juiced \\Predicted: Creamy}
            \label{qual7}
        \end{subfigure}
        \begin{subfigure}[c]{0.24\textwidth}
            \centering
            \includegraphics[width=0.9\textwidth]{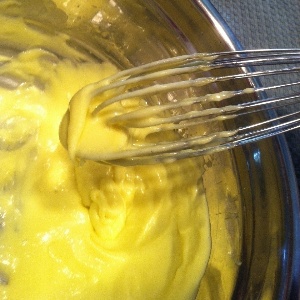}
            \caption{Label: Creamy\\\hspace{\textwidth}}
            \label{qual8}
        \end{subfigure}
        \begin{subfigure}[c]{0.24\textwidth}
            \centering
            \includegraphics[width=0.9\textwidth]{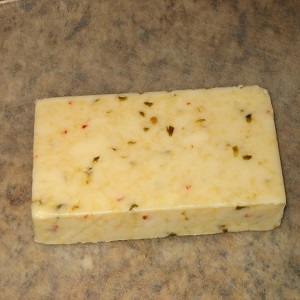}
            \caption{Label: Whole \\Predicted: Sliced}
            \label{qual9}
        \end{subfigure}
        \begin{subfigure}[c]{0.24\textwidth}
            \centering
            \includegraphics[width=0.9\textwidth]{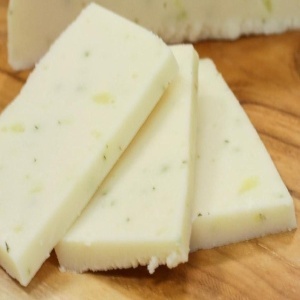}
             \caption{Label: Sliced\\\hspace{\textwidth}}
            \label{qual10}
        \end{subfigure}
        \caption{Misclassified sample with visually similar samples of the predicted class}
        \label{mislabel}
\end{figure}

However, the model failed to correctly classify a few samples.
According to the confusion matrix shown in \figureautorefname~\ref{confusion}, the model frequently misclassified the `juiced' and `whole' categories to the `creamy paste' and `sliced' respectively. Upon reviewing these samples, we observed that this could be due to images of the wrongly predicted classes being visually similar to the misclassified samples, as shown in \figureautorefname~\ref{mislabel}. In the training set of the `creamy' and `sliced' classes, there are several images, such as \figureautorefname~\ref{qual8} and \ref{qual10}, that are analogous to \figureautorefname~\ref{qual7} and \ref{qual9} respectively. Since the model learns to classify these images as belonging to the classes `creamy' and `sliced' during training, it is expected that similar images from other classes of the test set will also be classified accordingly.

\section{Conclusion and Future Work}\label{sec:conclusion}
Accurate classification of cooking objects is crucial for correctly handling food items by kitchen robots. In this regard, we have utilized the self-attention mechanism of the Vision Transformer architecture to contextualize information globally from images, along with leveraging the weights of a larger dataset.
This global attention and transfer learning allow the model to overcome the intra-class variations 
and the lack of inductive bias, respectively. As seen from the experiments, the proposed approach drastically outperforms the state-of-the-art works on the Cooking State Recognition Challenge Dataset. We also performed comprehensive experiments using several ablations to analyze the components of ViT thoroughly and compared it to baseline architectures to emphasize its significance. Furthermore, we illustrated the model's capability through visualization of the attention map. The structural similarity between some classes in the dataset was the basis of some misclassified samples, which is a limitation of our work. In the future,  Generative Adversarial Networks (GANs) can be used to augment the dataset and generate new training samples which may result in an even better performance.

\bibliographystyle{IEEEtran}
\bibliography{References}

\end{document}